\documentclass[a4paper,conference]{IEEEtran}
\IEEEoverridecommandlockouts
\usepackage[table,xcdraw]{xcolor} 
\usepackage{multirow} 
\usepackage{amsmath} 
\usepackage[raster,most]{tcolorbox}
\usepackage{xspace}               
\newcommand{\methodname}{FAVOR\xspace}  

\usepackage[utf8]{inputenc} 

\usepackage{tcolorbox}
\usepackage{xspace}

\usepackage{cite}
\usepackage{amsmath,amssymb,amsfonts}
\usepackage{algorithmic}
\usepackage{graphicx}
\usepackage{textcomp}
\usepackage{xcolor}
\usepackage{booktabs}
\usepackage{times}
\def\BibTeX{{\rm B\kern-.05em{\sc i\kern-.025em b}\kern-.08em
    T\kern-.1667em\lower.7ex\hbox{E}\kern-.125emX}}

\definecolor{ForestGreen}{rgb}{0, 0.69, 0.31}
\definecolor{NavyBlue}{rgb}{0, 0.44, 0.75}
\newcommand{\hgreen}[1]{\textcolor{ForestGreen}{\textbf{#1}}} 

\renewcommand{\methodname}{FAVOR\xspace}

\begin{document}
\title{Few-shot Vision-based Human Activity Recognition with MLLM-based Visual Reinforcement Learning\\

}


\author{
    \IEEEauthorblockN{
    Wenqi Zheng\textsuperscript{*},
    Yutaka Arakawa\textsuperscript{*}
    }
    \IEEEauthorblockA{
    \textsuperscript{*)}\textit{Graduate School and Faculty of Information Science and Electrical Engineering, Kyushu University, JAPAN}\\
    }
    \IEEEauthorblockA{E-mail: zheng.wenqi.005@s.kyushu-u.ac.jp, arakawa@ait.kyushu-u.ac.jp
    }
}

\maketitle

\begin{abstract}
Reinforcement learning in Large Reasoning Models enables learning from feedback on their outputs, making it particularly valuable in scenarios where fine-tuning data is limited. However, its application in multi-modal human activity recognition (HAR) domains remains largely underexplored. Our work extends the reinforcement learning to the human activity recognition domain with multimodal large language models. By incorporating the visual reinforcement learning in the training process, the model’s generalization ability on few-shot recognition can be greatly improved. Besides, the visual reinforcement learning can enhance the model reasoning ability and enable the explainable analysis in the inference stage. We name our Few-shot human Activity recognition method with Visual reinfORcement learning as FAVOR. Specifically, our approach firstly utilizes a multimodal large language model (MLLM) to generate multiple candidate responses for the human activity image, each containing reasoning traces and final answers. These responses are then evaluated using reward functions and the MLLM model is subsequently optimized using the Group Relative Policy Optimization (GRPO) algorithm. In this way, the MLLM model can be adapted to human activity recognition with only a few samples. Extensive experiments on 4 human activity recognition datasets and 5 different settings demonstrate the superiority of the proposed method.
\end{abstract}


\section{Introduction}
Human Activity Recognition (HAR) focuses on identifying and classifying human activities by analyzing data collected from various devices, including visual technologies and IoT-based systems~\cite{yin2024systematic,sun2022human,suda2023user}. Currently, two primary types of HAR systems are widely used: video-based systems and sensor-based systems~\cite{suda2023user}. Video-based systems rely on visual modalities such as RGB videos, skeleton data, and depth information, while sensor-based systems utilize devices like gyroscopes and accelerometers. Among these, video-based approaches dominate current HAR research due to their ability to capture richer information and provide contextual understanding of the scene. At present, HAR has been applied in a wide range of domains, including smart homes, healthcare, social sciences, rehabilitation engineering, fitness, and more~\cite{jin2024efficient,qi2024review,bastwesy2023tracking}. HAR has significantly enhanced human safety and well-being worldwide. This work primarily discusses vision-based HAR in daily-life scenarios.

\begin{figure}[t!]   
	\centering
	\includegraphics[width=\linewidth,scale=1.00]{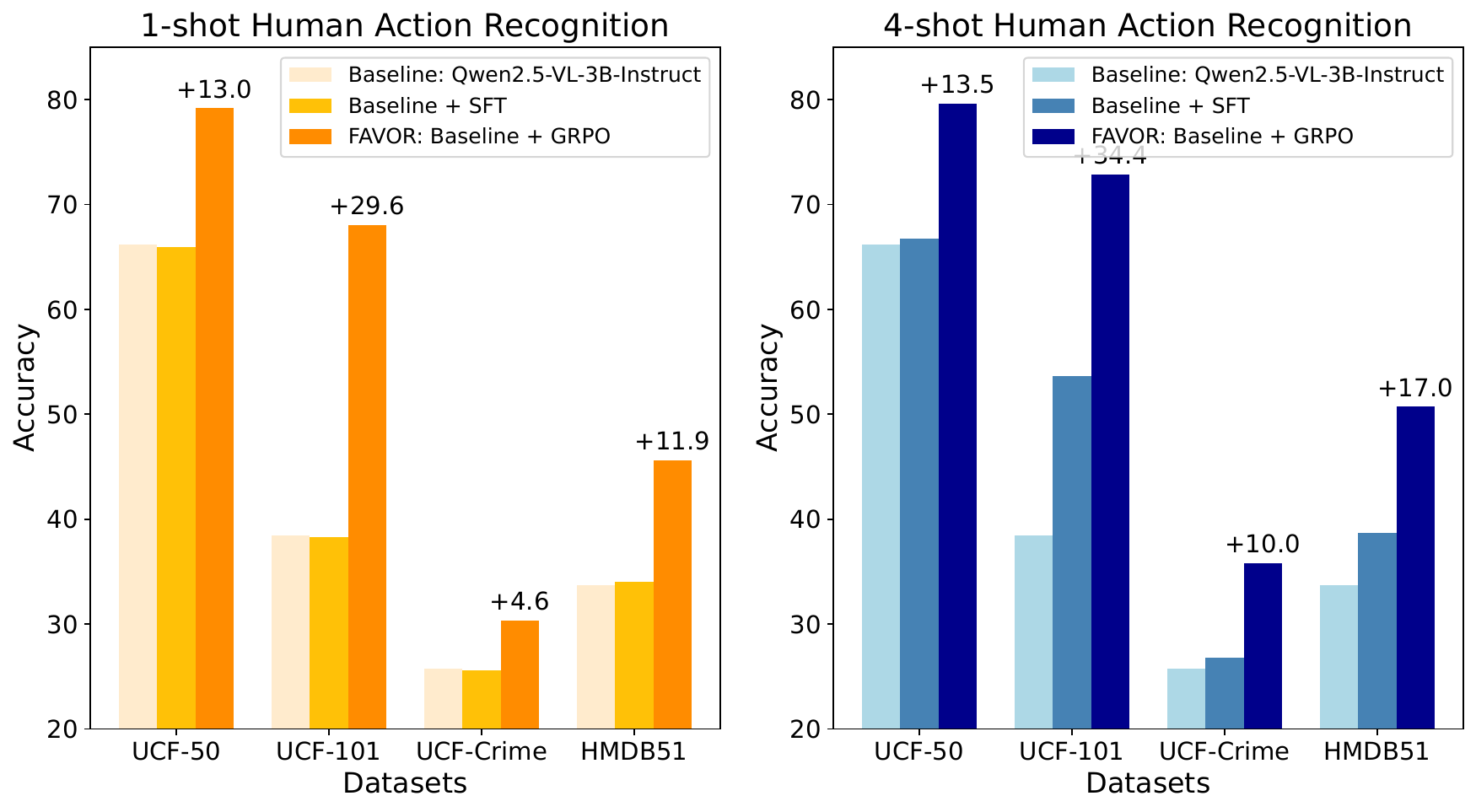}
	\caption{FAVOR outperforms traditional supervised fine-tuning (SFT) across various tasks in HAR classification.}
        \label{FigurOne}
        \vspace{-0.3cm}
\end{figure}

\begin{figure*}[ht!]   
	\centering
	\includegraphics[width=\linewidth,scale=1.00]{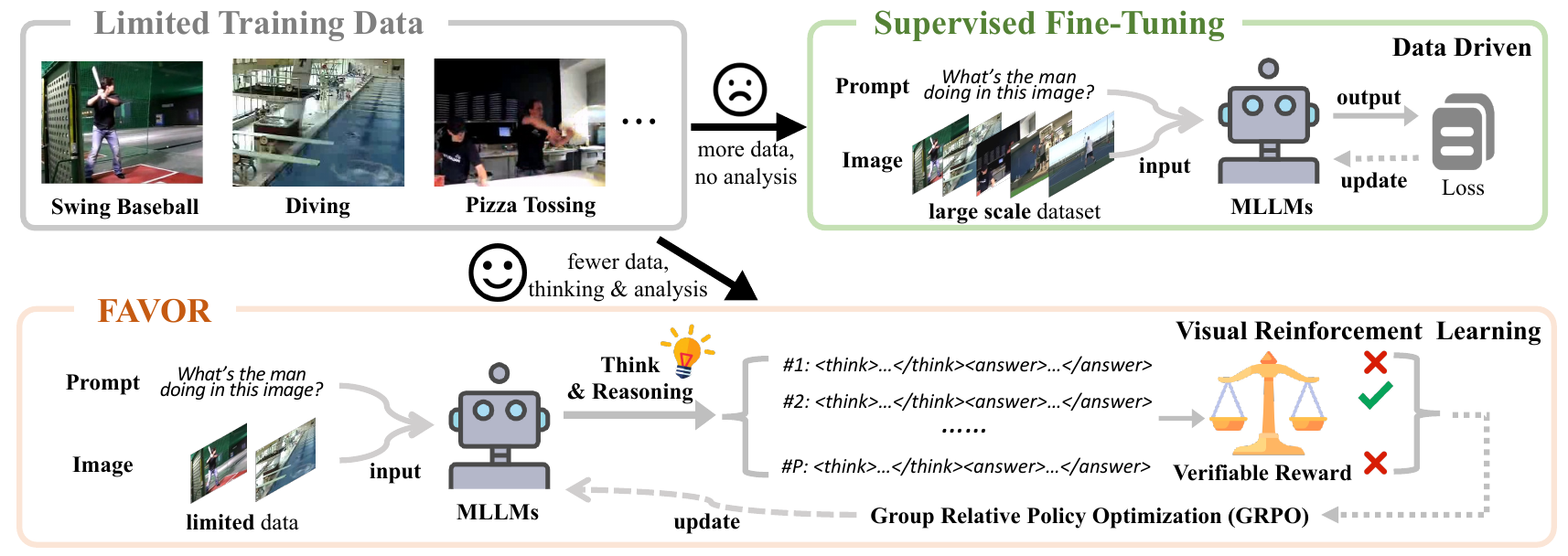}
    
    \vspace{-0.3cm}
	\caption{Overview of Human Activity Recognition (HAR) using Multimodal Large Language Models (MLLMs)}
        \label{FigureTwo}
        \vspace{-0.3cm}
\end{figure*}
Traditional HAR systems employ classical machine learning classifiers~\cite{yin2024systematic,sun2022human} to process manually extracted feature vectors from human activity data. The advent of deep learning has shifted this paradigm toward automated feature extraction. Neural architectures can directly process raw human activity data streams in an end-to-end manner~\cite{jin2024efficient,qi2024review}. Subsequently, the introduction of the transformer architecture further revolutionized deep learning~\cite{guo2025deepseek,zhou2024ristra,zhou2024dynamic}. By employing multi-head self-attention mechanisms, transformer models learn diverse representations from multiple perspectives through the collaboration of attention heads~\cite{yin2024systematic}. This approach outperforms traditional methods while often requiring fewer computational resources. However, most of these methods depend on large-scale labeled datasets and show limited generalization to previously unseen human activities~\cite{jin2024efficient}. Additionally, they primarily produce direct activity classification outputs and lack interpretability, which hinders their deployment in real-world scenarios. Besides, in the domain of HAR, their full potential and limitations remain largely unexplored, particularly in leveraging the rich diversity of multi-modal data~\cite{kumar2025llm}. 

Recently, many researchers have explored using Multimodal Large Language Models (MLLMs) to address HAR tasks~\cite{shen2025vlm,liu2025visual,li2024surveying,zhao2025r1,yin2024systematic,qi2024review,sun2022human}. While MLLMs generalize well after large-scale pretraining, Fine-Tuning further enhances their performance in HAR tasks. The Supervised Fine-Tuning (SFT) paradigm directly imitates the “ground truth” answers provided in high-quality datasets, thus relying on large amounts of training data. Another notable post-training method is Reinforcement Fine-Tuning (RFT), in which the reward score is determined by predefined rules. These rules evaluate the model’s responses and guide optimization based on feedback. A key distinction between RFT and SFT lies in data efficiency: RFT can effectively fine-tune a model using as few as dozens to thousands of samples, whereas SFT typically requires more data. In this paper, we introduce a few-shot human activity recognition method with visual reinforcement learning (FAVOR), which enables zero-shot and few-shot generalization to new tasks without domain-specific optimization. As a result, this approach empowers MLLMs to handle a wide range of multi-modal HAR tasks (see Fig~\ref{FigureTwo}). However, as illustrated in Fig~\ref{FigureTwo}, SFT directly produces outputs without multi-step inference or contextual reasoning, which often results in inaccurate responses~\cite{kumar2025llm,naveed2023comprehensive}. To address this limitation, we present the implementation details of FAVOR in Fig~\ref{FigureTwo}. Specifically, for each input, FAVOR leverages an MLLMs to generate multiple responses that contain both reasoning tokens and final answers. Importantly, we define rule-based, verifiable reward functions to guide policy optimization using Group Relative Policy Optimization (GRPO). For example, we design a tailored reward function for classification-based HAR tasks. By doing so, FAVOR is able to explore various possible reasoning paths and learn to optimize toward desired outcomes defined by these verifiable reward functions. Consequently, our approach shifts the training paradigm from data scaling in SFT to the strategic design of flexible reward functions customized for specific multi-modal HAR tasks.

Moreover, as shown in Fig~\ref{FigurOne}, in few-shot experiments, FAVOR demonstrates exceptional performance with minimal training data, showcasing significantly stronger few-shot learning capabilities compared to SFT. These diverse visual HAR tasks further underscore the crucial role of reinforcement learning in enhancing both visual perception and reasoning within multi-modal contexts. In summary, our key contributions are as follows:

\begin{itemize}
    \item We present FAVOR, a human activity recognition method that introduces visual reinforcement learning with verifiable rewards to the human activity recognition field, which proves to be effective even with scarce training data.
    \item A verifiable reward function is proposed for the human activity recognition task, which improves the recognition accuracy and enables explainable analysis.
    \item Extensive experiments on 4 human activity recognition datasets and 5 different settings demonstrate the superiority of the proposed method compared with baseline and supervised fine-tuning.
\end{itemize}

\section{Related Works}

\begin{figure*}[th!]   
	\centering
	\includegraphics[width=\linewidth,scale=1.00]{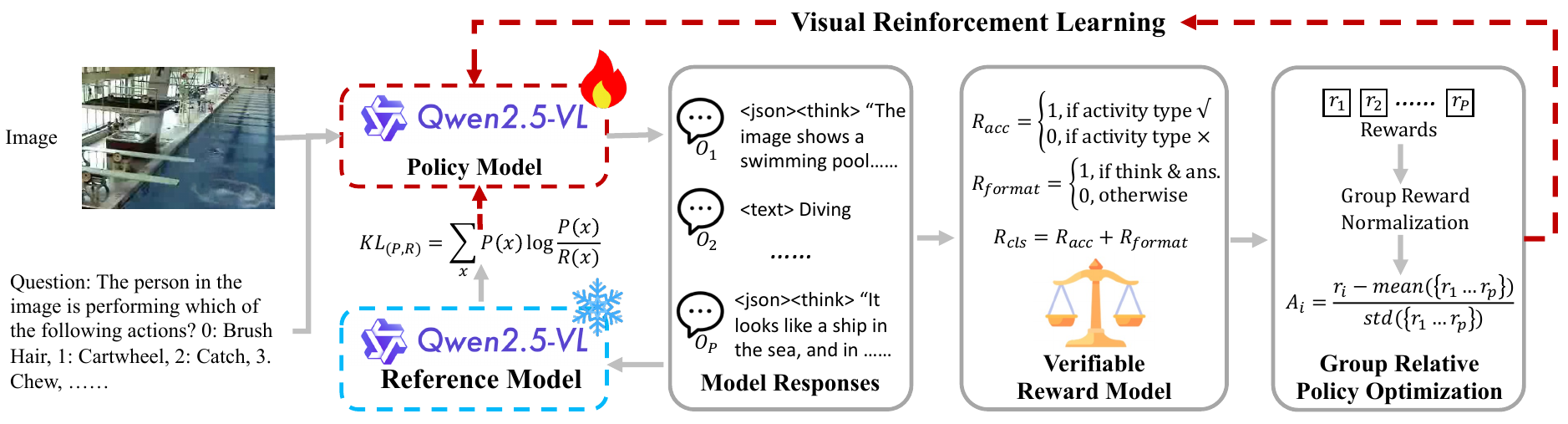}
	\caption{Framework of FAVOR}
        \label{FigureThree}
        \vspace{-0.3cm}
\end{figure*}
\subsection{Reinforcement Learning}

Multimodal Large Language Models (MLLMs) integrate multiple data modalities-such as audio, images, and text—into a unified framework~\cite{li2024surveying,naveed2023comprehensive,jin2024efficient}. In this context, researchers have demonstrated that the post-training phase can effectively improve a model's response quality and reasoning abilities, without relying heavily on extensive supervised data or complex prompting techniques~\cite{kumar2025llm}. Building upon these findings, recent research has increasingly focused on enhancing MLLMs' reasoning capabilities through reinforcement learning (RL) techniques. Specifically, RL enables models to make decisions by interacting with an environment and receiving rewards as feedback to maximize long-term gains. To further advance alignment and reasoning, Group Relative Policy Optimization (GRPO) was developed, which reformulates the alignment task as a ranking loss based on human-labeled preferences. Unlike traditional supervised methods, this approach directly optimizes the policy using group-based reward comparisons, while maintaining alignment with human intent~\cite{kumar2025llm}.

\subsection{Apply RL in HAR tasks}
Initially, reinforcement learning (RL) was applied to deep learning models—for example, by combining deep learning and RL to predict human activities in smart homes, or to control home robots for improved understanding of human interactions. Subsequently, with the rise of large language models (LLMs), RL has become essential for post-training. In particular, Reinforcement Learning with Human Feedback (RLHF) plays a critical role in aligning models with human intent. Moreover, reward models have proven effective in enhancing reasoning performance during inference. As research continues to advance, RL is now being increasingly applied to Multimodal Large Language Models (MLLMs). Notably, in the context of HAR, RL contributes to achieving better reasoning, alignment, and generalization across diverse multimodal HAR tasks. 

VisualThinker-R1-Zero~\cite{deng2025openvlthinker} demonstrated that applying R1 to a base MLLM could lead to significant performance improvements and trigger the emergence of the so-called “aha moment.” OpenVLThinker adopted an iterative self-improvement training strategy by alternating between supervised fine-tuning (SFT) and reinforcement learning (RL). Specifically, SFT was used to guide the model in acquiring an initial reasoning structure, while RL was employed to enhance its performance and generalization capabilities. Furthermore, VLM-R1~\cite{shen2025vlm} proposed a cross-modal reasoning pipeline and applied reinforcement learning to visual perception tasks. Similarly, R1-Omni~\cite{zhao2025r1} applied Reinforcement Learning with Verifiable Reward (RLVR) to an omni-multimodal large language model for emotion recognition. Open-Reasoner-Zero~\cite{hu2025open} demonstrated that even a simple rule-based reward strategy could improve both response length and benchmark performance. Notably, Visual-RFT~\cite{liu2025visual} enabled an MLLM to outperform models trained solely through SFT in object detection tasks. In contrast to these models, our approach employs a different base model and is evaluated in a distinct application setting.

\section{Methodology}
\subsection{Preliminary}

\subsubsection{Policy Gradient Optimization via Reinforce}

The reinforce algorithm is a policy gradient method used to improve decision-making by adjusting a model’s strategy (policy) based on the rewards received from its actions. It works by refining the probability distribution over possible actions, thereby increasing the likelihood of selecting actions that lead to better outcomes. At each iteration, the model updates its parameters ($\theta$) based on the performance of its past decisions, gradually optimizing its behavior over time. For a given question \( Q \), GRPO generates \( P \) distinct responses \( \{O_1, O_2, \ldots, O_P\} \) using the current policy \( \pi_{\theta_{\text{old}}} \). Each response is then evaluated to obtain a corresponding reward \( \{r_1, r_2, \ldots, r_P\} \). To assess their relative quality, GRPO normalizes the rewards by computing the mean and standard deviation:


 \[
A_i = \frac{r_i - \text{mean}(\{r_1, \ldots, r_P\})}{\text{std}(\{r_1, \ldots, r_P\})},
\tag{4}
 \]

Here, \( A_i \) denotes the normalized score for the \( i \)-th response. By normalizing the rewards, GRPO encourages the model to prioritize higher-quality responses within each group, helping it effectively distinguish between strong and weak outputs.




\subsubsection{Verifiable Rewards via Reinforce}
verifiable rewards can improve MLLMs on tasks with objectively correct answers, which directly employs a verification function to assess output correctness. Given an image and an input question \( Q \), the multimodal large language model \( M_{\theta} \) generates a series of responses, denoted as \( O \), which are evaluated by a verifiable reward function \( R(Q, O) \). This function assigns a score based on whether the output matches the ground truth:

\begin{equation}
R(Q, O) = 
\begin{cases}
1, & \text{if } O = \text{ground truth}, \\
0, & \text{otherwise}.
\end{cases}
\tag{1}
\end{equation}

The optimization objective is defined as:

\begin{equation}
\max_{\pi_{\theta}} \mathbb{E}_{O \sim \pi_{\theta}(Q)} \left[ R_{\text{RLVR}}(Q, O) \right],
\tag{2}
\end{equation}

where

\begin{equation}
R_{\text{RLVR}}(Q, O) = R(Q, O) - \alpha \cdot \text{KL} \left[ \pi_{\theta}(O \mid Q) \| \pi_{\text{ref}}(O \mid Q) \right].
\tag{3}
\end{equation}

Here, \( \pi_{\text{ref}} \) denotes the reference model prior to optimization, and \( \alpha \) is a hyperparameter that balances reward maximization against KL-divergence regularization, encouraging the optimized model to remain close to the original. In this work, we integrate GRPO with multimodal large language models to leverage the strengths of both approaches, enhancing the model's capabilities in reasoning and activity recognition.


\begin{figure}
\begin{tcolorbox}[title=Trivial Reasoning Trajectory, label=meaningless_response]
\textbf{Question:} \texttt{<image>} The person in the image is performing which of the following activities? 0: brush\_hair, 1: cartwheel, 2: catch, \texttt{*****} \newline
\textbf{Response:}  First output the thinking process in \texttt{<think>} \texttt{</think>} tags and then output the final answer in \texttt{<answer>} \texttt{</answer>} tags. Output the final answer in JSON format.
\end{tcolorbox}
\vspace{-3mm}
\caption{Example response of applying GRPO to base models.}
\label{Figurefive}
\vspace{-3mm}
\end{figure}

\subsection{FAVOR}
\subsubsection{FAVOR framework}
The FAVOR framework, illustrated in Fig.~\ref{FigureThree}, utilizes Qwen2.5-VL-3B-Instruct~\cite{bai2025qwen2} as the policy model \( \mathcal{\pi}_{\theta} \), which takes multimodal inputs from the user—namely images and questions—and generates a reasoning process along with a set of candidate responses. We design a prompt format that guides the model to generate a reasoning trace before producing the final answer, as shown in Fig.~\ref{Figurefive}. During training, the reward function encourages the model to produce structured outputs that include both a reasoning process and a final answer. The reasoning trace facilitates self-improvement during reinforcement fine-tuning, while the answer-based reward drives policy optimization. By combining two rewards, the model is guided to generate predictions that are both accurate and well-structured. Each response is evaluated through a verifiable reward function, and group-wise reward computation is performed to assess the quality of each output. These evaluations are then used to update the policy model. To ensure stable training, FAVOR incorporates KL-divergence regularization to constrain the deviation between the current policy model \( \mathcal{\pi}_{\theta} \) and a reference model \( \pi_{\theta_{\text{old}}} \).

\subsubsection{Verifiable Reward in HAR tasks}

The reward model plays a crucial role in RL by checking whether the model's prediction exactly matches the ground-truth answer. This work focuses on reasoning over images to classify the activity performed by the person, guided by a given question. So, the reward function \( R_{\text{cls}} \) consists of two components: an accuracy reward \( R_{\text{acc}} \) and a format reward \( R_{\text{format}} \). The overall reward is defined as follows:

\[
R_{\text{cls}} = R_{\text{acc}} + R_{\text{format}}. \tag{5}
\]

Our reinforcement learning strategy avoids complex reward modeling, opting instead for a simple rule-based reward function that encourages responses with correct content and proper formatting:

\begin{table}[ht!]
\centering
\caption{Hyperparameters used in FAVOR training.}
\label{TableTwo}
\begin{tabular}{@{}lc@{}}
\toprule
\textbf{Setting} & \textbf{Value} \\
\midrule
Batch size per device              & 8 \\
Data\_seed                          &4\\
Gradient accumulation steps        & 2 \\
Training steps                     & 20 \\
Learning rate                      & $5 \times 10^{-5}$ \\
Temperature                        & 1.0 \\
Maximum response length            & 2048 \\
Responses per GRPO step            & 16 \\
KL coefficient                     & 0.04 \\
\bottomrule
\end{tabular}
\vspace{-3mm}
\end{table}

\begin{itemize}
    \item \textbf{Accuracy reward (+1):} Assigned when the response provides a correct final answer.
    \item \textbf{Format reward (+1):} Assigned when the response wraps its reasoning in \texttt{<think>} tags and the final answer in \texttt{<answer>} tags.
    \item \textbf{Otherwise:} A reward of 0 is assigned.
\end{itemize}

\section{Experiments}
\subsection{Datasets}

Most state-of-the-art activity recognition methods are evaluated on datasets featuring limited activity categories and captured in controlled environments. In contrast, the UCF50~\cite{reddy2013recognizing} dataset consists of real-world YouTube videos spanning 50 diverse activity categories. The HMDB51~\cite{kuehne2011hmdb} dataset includes 51 categories, encompassing both facial expressions (e.g., smiling, laughing, chewing, and talking) and 47 full-body activities. UCF101~\cite{soomro2012ucf101} offers 101 human activity classes with over 13,000 video clips totaling 27 hours of footage. The UCF-Crime~\cite{sultani2018real} dataset comprises 1,900 long, untrimmed surveillance videos depicting 13 types of realistic anomalies—such as fighting, road accidents, burglary, and robbery—as well as normal daily activities. These videos are recorded under realistic conditions, often involving camera motion and cluttered backgrounds, which makes this dataset particularly challenging for activity recognition. Many existing approaches struggle on these datasets, which contain videos collected from the web and often present challenges such as camera motion, poor lighting, cluttered scenes, varying scales and viewpoints, and inconsistent focus on the human activity itself. In this study, we demonstrate that our FAVOR model is capable of handling HAR reasoning and classification tasks. We train our model on four datasets: UCF50~\cite{reddy2013recognizing}, UCF101~\cite{soomro2012ucf101}, UCF-Crime~\cite{sultani2018real}, and HMDB51~\cite{kuehne2011hmdb}.

\subsection{Evaluation}

\begin{table*}[t]
\caption{Few-shot Recognition Accuracy on HAR datasets (1-shot, 2-shot and 4-shot).}
\vspace{-6mm}
\label{TableTransposed_Part1}
\begin{center}
\scalebox{0.92}{
\begin{tabular}{l|c|cccccccccccccc}
\toprule
\multirow{2}{*}{\bf Dataset} & \multirow{2}{*}{\bf Baseline} & \multicolumn{4}{c}{\textit{\bf 1-shot}} & \multicolumn{4}{c}{\textit{\bf 2-shot}} & \multicolumn{4}{c}{\textit{\bf 4-shot}} \\
\cmidrule(lr){3-6} \cmidrule(lr){7-10} \cmidrule(lr){11-14}
 & & +SFT & $\Delta$SFT & +FAVOR & $\Delta$FAVOR & +SFT & $\Delta$SFT & +FAVOR & $\Delta$FAVOR & +SFT & $\Delta$SFT & +FAVOR & $\Delta$FAVOR \\
\midrule
Average & 41.00 & 40.96 & -0.04 & 55.78 & \hgreen{+14.78} & 42.51 & +1.51 & 56.32 & \hgreen{+15.32} & 46.47 & +5.47 & 59.75 & \hgreen{+18.75} \\
UCF-50~\cite{reddy2013recognizing}  & 66.16 & 65.96 & -0.20 & 79.17 & \hgreen{+13.01} & 67.06 & +0.90 & 80.14 & \hgreen{+13.98} & 66.78 & +0.62 & 79.65 & \hgreen{+13.49} \\
UCF-101~\cite{soomro2012ucf101} & 38.41 & 38.31 & -0.10 & 68.03 & \hgreen{+29.62} & 42.94 & +4.53 & 72.02 & \hgreen{+33.61} & 53.65 & +15.24 & 72.84 & \hgreen{+34.43} \\
UCF-Crime~\cite{sultani2018real} & 25.77 & 25.56 & -0.21 & 30.32 & \hgreen{+4.55} & 25.82 & +0.05 & 26.75 & \hgreen{+0.98} & 26.80 & +1.03 & 35.81 & \hgreen{+10.04} \\
HMDB51~\cite{kuehne2011hmdb} & 33.67 & 34.02 & +0.35 & 45.59 & \hgreen{+11.92} & 34.20 & +0.53 & 46.38 & \hgreen{+12.71} & 38.66 & +4.99 & 50.71 & \hgreen{+17.04} \\
\bottomrule
\end{tabular}
}
\end{center}
\vspace{-3mm}
\end{table*}

\begin{table}[t]
\caption{Few-shot Recognition Accuracy on HAR datasets.}
\vspace{-3mm}
\label{TableTransposed_Part2_Formatted}
\begin{center}
\scriptsize
\resizebox{\linewidth}{!}{
\begin{tabular}{l|c|cccc}
\toprule
\textbf{Models} & \textbf{Avg.} & UCF-50~\cite{reddy2013recognizing} & UCF-101~\cite{soomro2012ucf101} & UCF-Crime~\cite{sultani2018real}  & HMDB51~\cite{kuehne2011hmdb} \\
\midrule
Baseline & 41.00 & 66.16 & 38.41 & 25.77 & 33.67 \\
\midrule
\multicolumn{6}{c}{\textbf{\textit{8-shot}}} \\
\midrule
$+$ SFT & 53.06 & 78.54 & 67.87 & 24.44 & 41.38 \\
$\Delta$SFT & +12.06 & +12.38 & +29.46 & -1.33 & +7.71 \\

$+$ \methodname & 61.22 & 80.94 & 75.27 & 36.82 & 51.83 \\
$\Delta$FAVOR & \hgreen{+20.22} & \hgreen{+14.78} & \hgreen{+36.86} & \hgreen{+11.05} & \hgreen{+18.16} \\
\midrule
\multicolumn{6}{c}{\textbf{\textit{16-shot}}} \\
\midrule
$+$ SFT & 58.63 & 82.93 & 75.79 & 27.84 & 47.96 \\
$\Delta$SFT & +17.63 & +16.77 & +37.38 & +2.07 & +14.29 \\

$+$ \methodname & 63.55 & 83.44 & 79.59 & 36.86 & 54.31 \\
$\Delta$FAVOR & \hgreen{+22.55} & \hgreen{+17.28} & \hgreen{+41.18} & \hgreen{+11.09} & \hgreen{+20.64} \\
\bottomrule
\end{tabular}
}
\end{center}
\vspace{-3mm}
\end{table}

\begin{figure}[th!]   
	\centering
	\includegraphics[width=\linewidth,scale=1.00]{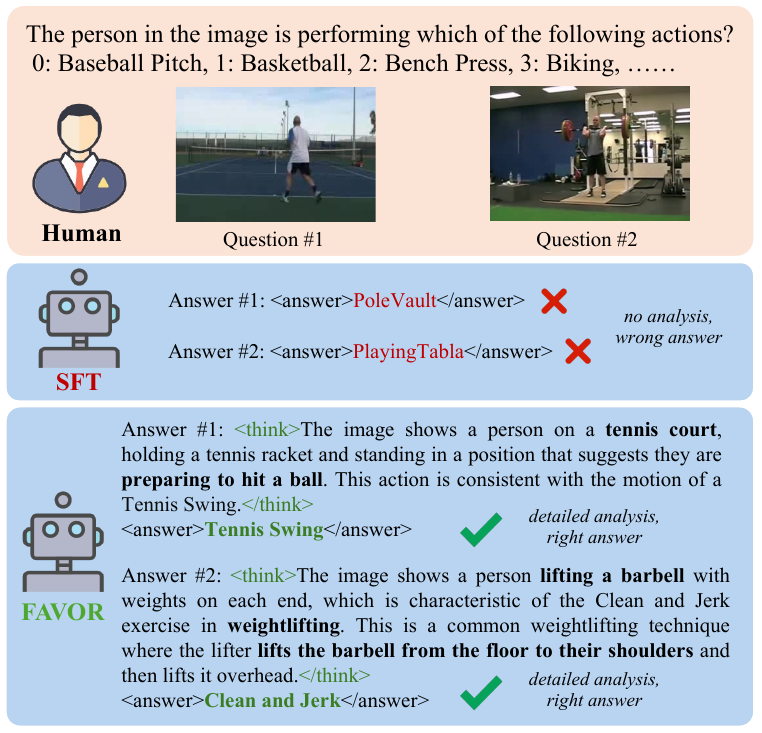}
        \vspace{-0.3cm}
	\caption{Results of HAR Image Classification}
        \label{Figurefour}
        \vspace{-0.3cm}
\end{figure}
To evaluate the reasoning and classification capabilities of our method, we conduct experiments on four datasets. We first obtain video from each datasets and randomly extract a single frame from several videos within each category to construct an new image dataset. These extracted frames are categorized according to their original video class across the datasets. Subsequently, a small number of images are randomly selected from each category to form the training set. The remaining images are used as the test set for subsequent evaluation and testing. We adopt a few-shot setting to assess the model’s recognition ability, applying reinforcement learning to limited data. Specifically, we train Qwen2.5-VL-3B-instruct using both reinforcement learning (RL) and supervised fine-tuning (SFT) on randomly sampled 1-shot, 2-shot, 4-shot, 8-shot, and 16-shot examples drawn from the training set. The remaining data is used as the test set for evaluation.

\subsection{Implementation Details}

The model is trained for 20 epochs using a learning rate of \(5\times10^{-5}\) and a temperature of 1.0. The maximum response length is set to 2048 tokens. During GRPO optimization, we sample 8 responses per step and use a KL coefficient of 0.04, as shown in Table~\ref{TableTwo}. To demonstrate the effectiveness of directly applying GRPO to the base model, we compare our method against two baselines: the non-SFT pre-trained model and the base model fine-tuned with SFT on the same dataset.



\subsection{Main Results}
We evaluate the few-shot recognition accuracy of various models on multiple HAR datasets, including UCF-50~\cite{reddy2013recognizing}, UCF-101~\cite{soomro2012ucf101}, UCF-Crime~\cite{sultani2018real}, and HMDB51~\cite{kuehne2011hmdb}. Our backbone model is the Qwen2.5-VL-3B-Instruct, which we fine-tune under different learning paradigms. As shown in Table~\ref{TableTransposed_Part1} and Table~\ref{TableTransposed_Part2_Formatted}, we compare three settings: Baseline, Supervised Fine-Tuning (SFT), and our proposed \methodname{} method.

Throughout various few-shot settings, \methodname{} consistently outperforms the SFT and baseline. In the low-shot regime, such as \textbf{1-shot}, \methodname{} achieves an average accuracy improvement of \textbf{+14.78\%} over the baseline, while SFT slightly underperforms (-0.04\%). The gains are especially prominent on UCF-101 (+29.62\%) and UCF-50 (+13.01\%). This trend continues under the \textbf{2-shot} and \textbf{4-shot} settings, where \methodname{} yields average gains of \textbf{+15.32\%} and \textbf{+18.75\%}, respectively. The performance boosts remain most notable on UCF-101 (up to +34.43\%) and HMDB51 (up to +17.04\%), demonstrating the model’s ability to generalize well from limited examples.

In higher-shot scenarios, the advantage of \methodname{} becomes even more pronounced. Under the \textbf{8-shot} setting, \methodname{} achieves an average accuracy of \textbf{61.22\%}, outperforming the Baseline (41.00\%) and SFT (53.06\%). It delivers consistent improvements across all datasets, including a significant \textbf{+11.05\%} gain over SFT on the challenging UCF-Crime dataset, where SFT otherwise suffers a slight performance drop (-1.33\%). In the \textbf{16-shot} setting, \methodname{} maintains its lead with an average accuracy of \textbf{63.55\%}, reflecting a \textbf{+22.55\%} improvement over Baseline and \textbf{+4.92\%} over SFT. Again, the highest gains appear on UCF-101 (\textbf{+41.18\%}) and HMDB51 (\textbf{+20.64\%}).

These results clearly indicate that \methodname{} significantly enhances few-shot recognition performance in HAR tasks. It not only outperforms conventional SFT in low-data regimes but also exhibits better scalability and robustness. Importantly, in some cases, SFT shows minimal or even negative improvements over the baseline, while \methodname{} delivers consistent and often large boosts—underscoring its promise as a generalizable solution for real-world activity recognition in data-scarce settings.

\subsection{Ablation }
To investigate the performance contributions of different components in our method, we conduct an ablation study on the UCF-50 2-shot dataset, as shown in Table~\ref{TableThree}. Starting from the base Qwen2.5-VL-3B-Instruct model, which achieves 66.16\% accuracy, we examine the impact of several key design choices. Notably, applying FAVOR with frozen vision modules leads to a significant improvement, reaching 79.69\% accuracy. This suggests that keeping the vision encoder fixed allows the model to better allocate its capacity toward learning effective reasoning strategies during reinforcement learning. Furthermore, we explore the effect of varying the number of sampled responses ($P$) during training. As $P$ increases from 2 to 16, the model's performance consistently improves—from 75.34\% at $P=2$ to 80.14\% at $P=16$. This highlights the benefit of richer trajectory exploration during optimization, allowing the model to better refine its policy over diverse reasoning paths. Overall, the results reveal that even without fine-tuning the vision backbone, substantial gains can be achieved by properly configuring the reinforcement learning setup. This underscores the importance of reward-driven reasoning exploration in multimodal learning and points to intricate dynamics between vision and language modules that merit further investigation.

\begin{table}[t]
\caption{Ablation results on the UCF-50 2-shot dataset.}
\vspace{-3mm}
\label{TableThree}
\begin{center}
\scalebox{0.95}{
\begin{tabular}{l|c}
\toprule
\textbf{Ablation Setup} & \textbf{Accuracy (\%)} \\
\midrule
Base model: Qwen2.5-VL-3B-Instruct & 66.16 \\
FAVOR with frozen vision modules & 79.69 \\
FAVOR with responses num $P=2$ & 75.34 \\
FAVOR with responses num $P=4$ & 78.32 \\
FAVOR with responses num $P=8$ & 79.93 \\
FAVOR: trainable vision modules, responses num $P=16$ & 80.14 \\
\bottomrule
\end{tabular}
}
\end{center}
\vspace{-3mm}
\end{table}

\section{Conclusion}
In this paper, we present FAVOR,  a method to adapt GRPO-based reinforcement learning for improving the model reasoning ability of MLLMs. FAVOR employs a verifiable reward system that reduces reliance on manual labeling and simplifies reward computation. It achieves strong performance across various HAR tasks. Extensive experiments demonstrate that FAVOR excels in reasoning and few-shot learning, consistently outperforming supervised fine-tuning (SFT) with minimal data. Our findings highlight the potential of visual reinforcement learning to enhance MLLMs for more efficient and effective HAR tasks.

\section*{Acknowledgment}
This work was partially supported by Hagiwara Foundation of Japan 5th Research Grant and the China Scholarship Council during the doctoral study at Kyushu University.


\bibliographystyle{plain}      
\bibliography{main}

\end{document}